\documentclass{article}
\PassOptionsToPackage{numbers,compress}{natbib}
\usepackage[dblblindworkshop, final]{neurips_2026}
\workshoptitle{Towards Test-Time Continual Learning Agents (TTCL)}

\usepackage{amsmath,amsfonts,bm}

\def\eqref#1{equation~\ref{#1}}

\def\1{\bm{1}}

\DeclareMathAlphabet{\mathsfit}{\encodingdefault}{\sfdefault}{m}{sl}
\SetMathAlphabet{\mathsfit}{bold}{\encodingdefault}{\sfdefault}{bx}{n}

\usepackage[T1]{fontenc}
\usepackage{subcaption}
\usepackage{hyperref}
\hypersetup{hidelinks}
\usepackage{url}
\usepackage{graphicx}
\usepackage{enumitem}
\usepackage{booktabs}
\usepackage{multirow}
\usepackage{microtype}
\usepackage{placeins}

\title{Learning What to Remember: Resource-Efficient Self-Improvement for Lifelong Agents}

\author{
Guannan Lai$^{1,2}$ \quad
Zi-Yang Bai$^{1,2}$ \quad
Jun-Peng Jiang$^{1,2}$
\quad
Han-Jia Ye$^{1,2}$
\\
{
$^{1}$School of Artificial Intelligence, Nanjing University
}\\
{
$^{2}$National Key Laboratory for Novel Software Technology, Nanjing University
}\\
{
\texttt{\{laign, jiangjp,  yehj\}@lamda.nju.edu.cn\quad baizy@smail.nju.edu.cn}
}
}

\begin{document}
\maketitle

\begin{abstract}
Self-improving agents aim to continuously improve their behavior by learning
from experience after deployment. However, accumulating and reusing more
experience does not necessarily lead to better decisions, particularly when
generation, interaction, and memory resources are jointly constrained.
Under such stringent constraints, we find that several existing lifelong-memory
methods provide limited benefits over simple retrieval or even no-memory
agents. This reveals a fundamental challenge for resource-efficient
self-improvement: agents must learn not only which past experiences are relevant,
but which are actually worth reusing to improve future decisions.
When memory capacity and opportunities for reasoning and recovery are limited,
retrieving an ineffective experience can consume scarce resources and hinder
downstream execution. To address this challenge, we propose
\textbf{TokenWorth}, a lightweight feedback-driven memory adaptation framework
that learns query-dependent memory worth from sparse execution feedback.
Without updating the backbone model, TokenWorth uses the learned worth to
guide both experience retrieval and bounded memory management, enabling
the agent to continually adapt how it reuses accumulated experience.
Across three LifelongAgentBench environments and both local and API-based
deployments, TokenWorth achieves the highest task success rate while using
the fewest output tokens among the evaluated methods under the specified
resource budgets. These results suggest that effective self-improvement
under resource constraints depends not on accumulating more experience,
but on learning what is worth remembering.
Our code is publicly available at
\url{https://github.com/AIGNLAI/TokenWorth}.
\end{abstract}

\section{Introduction}

Self-improving agents continually accumulate and reuse experience from
evolving task streams to improve their behavior after deployment
\citep{zheng2026lifelong}. Such experience-driven adaptation is particularly
valuable for long-running agents facing evolving user needs,
environments, and tasks \citep{cheng2024sociodojo,fang2025comprehensive,luo2025agrail}.
However, accumulating more experience does not necessarily translate into
better decisions. In practice, generation is constrained by latency,
computation, or inference cost; interaction opportunities are finite; and
only a bounded amount of historical experience can be stored or reused
\citep{gupta2025ai,kwon2023efficient,yuan2026large}.
This raises a fundamental question: \emph{how can agents effectively
improve from experience when generation, interaction, and memory resources
are jointly constrained?}

\begin{figure*}[t]
    \centering
    \begin{subfigure}[t]{0.36\textwidth}
        \centering
        \includegraphics[width=\linewidth]{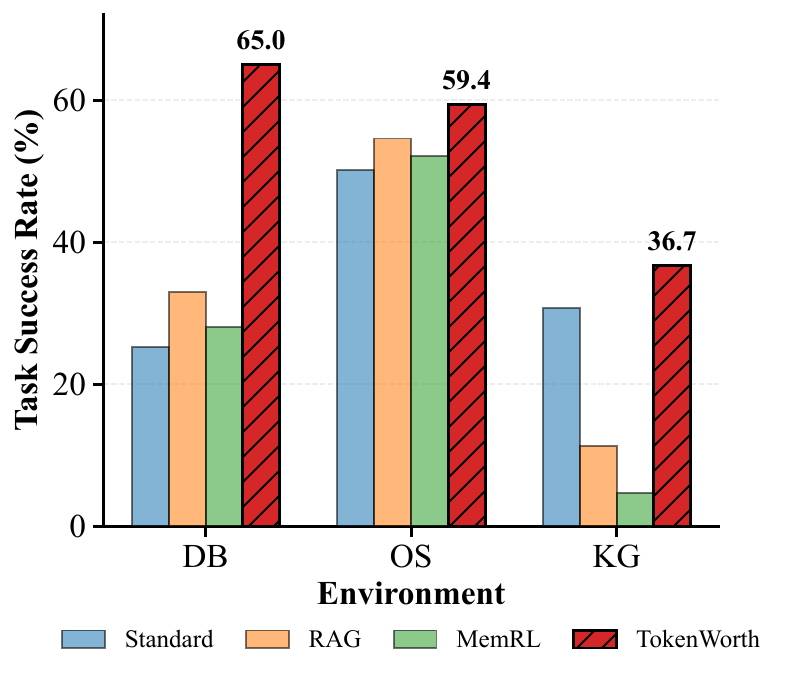}
        \label{fig:intro-collapse}
    \end{subfigure}
    \hfill
    \begin{subfigure}[t]{0.62\textwidth}
        \centering
        \includegraphics[width=\linewidth]{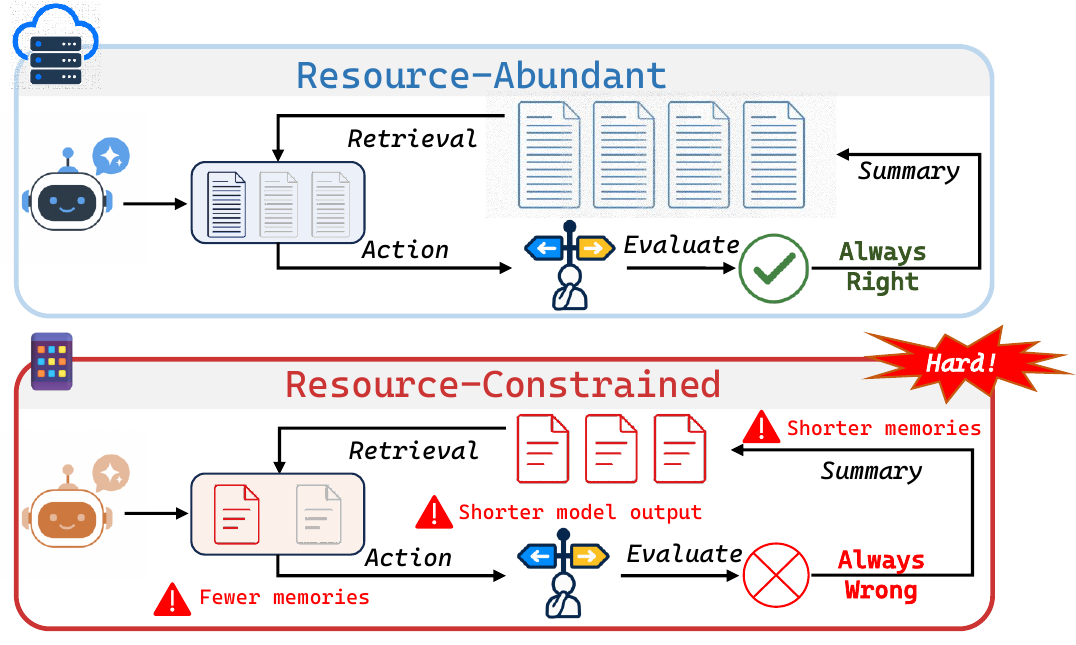}
        \label{fig:intro-setting}
    \end{subfigure}
    \caption{
    Left: Representative local results with Qwen2.5-14B-Instruct under the
    evaluated stringent joint resource constraints; TokenWorth remains effective
    where several memory baselines do not.
    Right: Resource-constrained lifelong agents adapt from experience under
    bounded generation, interaction, and memory budgets.
    }
    \vspace{-0.5cm}
    \label{fig:intro}
\end{figure*}

Our evaluation reveals that existing memory mechanisms often struggle under
such constraints. As shown in Figure~\ref{fig:intro}(a), several sophisticated
lifelong-memory approaches provide limited improvement over simple retrieval
or even no-memory agents, and can substantially underperform in some
environments. These results suggest that accumulating and reusing experience
alone is insufficient for reliable adaptation under stringent resource
constraints. Importantly, this limitation cannot be attributed simply to
insufficient memory capacity: under the same resource budget, different
memory policies can lead to markedly different downstream task performance.

To study this challenge systematically, we formalize
\emph{resource-constrained lifelong agents}: agents that must continually
adapt from experience while operating under bounded generation, interaction,
and memory budgets. As illustrated in Figure~\ref{fig:intro}(b), we jointly
constrain the resources available for task execution and experience reuse.
Specifically, the agent operates with a bounded output budget and limited
interaction rounds, while its external memory is constrained in capacity,
retrieval count, and injection length. The backbone remains frozen throughout
the task stream, allowing us to study experience-driven adaptation under
a controlled, stringent resource budget.

Why does experience-based adaptation become fragile under such constraints?
When only a small amount of experience can be retained or reused, and
opportunities for reasoning and recovery are limited, each memory decision
becomes more consequential. Semantic relevance alone may be insufficient:
an experience can appear relevant to the current query yet provide little
downstream utility. Conversely, useful experience that is not retained or
selected cannot be recovered by simply injecting more context or extending
the interaction. Resource scarcity therefore exposes a fundamental distinction
between accumulating experience and effectively learning from it. The central
challenge is not merely to retrieve relevant experience, but to determine
which experiences are actually \emph{worth} retaining and reusing to improve
future decisions.

To address this challenge, we propose \textbf{TokenWorth}, a lightweight
feedback-driven memory adaptation framework that learns query-dependent
memory worth from sparse execution feedback. TokenWorth combines an
interpretable online base scorer with a residual Recursive Feature Machine
(RFM), allowing stable global patterns of memory usefulness to be
complemented by nonlinear, query-specific corrections. The learned worth
guides both memory retrieval and bounded memory-bank management, enabling
the agent to continually adapt its experience-reuse policy without
backbone updates or additional language-model calls.

We evaluate TokenWorth on three LifelongAgentBench environments covering
database interaction, operating-system tasks, and knowledge-graph reasoning,
under both local and API-based deployments. Across all six
environment--deployment combinations, TokenWorth achieves the highest
task success rate among the evaluated methods while using the fewest output
tokens. These results suggest that resource-efficient adaptation depends
not on accumulating more experience, but on learning what is worth
remembering.

Our contributions are threefold:
\begin{itemize}[leftmargin=*]
    \item We formulate \emph{resource-constrained lifelong agents} and
    investigate experience-driven adaptation under stringent joint
    constraints on generation, interaction, and memory.

    \item We propose \textbf{TokenWorth}, a lightweight online framework
    that learns query-dependent memory worth from execution feedback
    and uses it consistently for memory retrieval and bounded
    memory management.

    \item We evaluate TokenWorth across three LifelongAgentBench
    environments under both local and API-based deployments,
    demonstrating improved task success and generation efficiency
    under stringent resource constraints.
\end{itemize}
\section{Related Work}

\paragraph{Experience-Driven Self-Improvement in Language Agents.}
Self-improving language agents aim to continually improve from accumulated
interaction experience, enabling adaptation after deployment without
necessarily updating backbone parameters
\citep{fang2025comprehensive,zheng2026lifelong}.
Early agent systems demonstrated that past trajectories, reflections, or
reusable skills can improve future decisions, including
Reflexion~\citep{shinn2023reflexion}, ExpeL~\citep{zhao2024expel},
Voyager~\citep{wangvoyager}, and SocioDojo~\citep{cheng2024sociodojo}.
More recent work has developed explicit memory and skill mechanisms
for continual adaptation, such as MemoryBank~\citep{zhong2024memorybank},
Mem0~\citep{chhikara2025mem0}, MemRL~\citep{zhang2026memrl},
MemSkill~\citep{zhang2026memskill}, and
AutoSkill~\citep{yang2026autoskill}.
Meanwhile, LifelongAgentBench~\citep{zheng2025lifelongagentbench},
AgentCL~\citep{shu2026agentcl},
SkillLearnBench~\citep{zhong2026skilllearnbench}, and
XSkill~\citep{jiang2026xskill} provide systematic evaluations of
experience-driven adaptation.
While these studies explore how agents accumulate and exploit experience,
we investigate how agents can learn which experiences are worth reusing
to support effective adaptation under stringent resource constraints.

\paragraph{Resource-Constrained Agent Adaptation.}
Resource efficiency has been extensively studied for LLM inference
and serving, including memory-efficient execution~\citep{kwon2023efficient}
and collaborative inference under computation and communication
constraints~\citep{yuan2026large}.
Recent work has extended resource considerations to agent adaptation.
\citet{hajimiri2026online} study whether online memory and skill modules
remain beneficial under fixed inference-token budgets, while
\citet{wu2026forget} investigate continual learning with hard-bounded
on-device memory.
BudgetMem~\citep{zhang2026learning} considers query-aware routing
across runtime memory budgets, and \citet{chen2026learning} model
agentic memory value using multiple factors for retention and retrieval.
These studies highlight the importance of resource-aware experience
reuse, but primarily focus on individual constraints, runtime budgeting,
or memory valuation in isolation.
In contrast, we study experience-driven self-improvement under
stringent joint constraints on generation, interaction, and memory.
We propose \textbf{TokenWorth}, which learns query-dependent memory
worth from online execution feedback to guide both retrieval and
bounded memory management, enabling continual adaptation without
backbone updates.
\section{Resource-Constrained Lifelong Agents}
\label{sec:problem_formulation}

We consider a lifelong agent that sequentially processes an online task stream
$\mathcal{Q}=\{q_1,q_2,\ldots,q_T\}$.
At step $t$, the agent receives task $q_t$ and may retrieve a set of past experiences
$\mathcal{R}_t$ from its current memory bank
$\mathcal{B}_t=\{m_1,\ldots,m_{N_t}\}$.
Conditioned on $q_t$ and $\mathcal{R}_t$, the backbone model interacts with the
environment for $J_t$ rounds, producing a trajectory
\[
\tau_t=\{(o_{t,j},a_{t,j})\}_{j=1}^{J_t},
\]
where $o_{t,j}$ and $a_{t,j}$ denote the observation and model output at
interaction round $j$, respectively.
The environment then returns a task-level reward $r_t\in\{0,1\}$.
After observing this outcome, the memory policy may update the bank from
$\mathcal{B}_t$ to $\mathcal{B}_{t+1}$.
We focus on a frozen-backbone setting, where experience-driven adaptation
occurs through online updates to external memory and its management policy,
rather than model parameters.

We impose resource constraints on both task execution and experience reuse.
For execution, the total amount of model generation associated with each task,
including both environment interaction and any method-specific auxiliary
language-model calls, is bounded by
\[
\sum_{j=1}^{J_t}|a_{t,j}|_{\mathrm{tok}} + A_t^{\mathrm{aux}}
\le O_{\max},
\qquad J_t \le J_{\max},
\]
where $A_t^{\mathrm{aux}}$ denotes the output tokens generated by auxiliary
language-model calls such as memory construction, reflection, or skill extraction.
For memory, the persistent bank is bounded by
\[
|\mathcal{B}_t| \leq B_{\max},
\]
while the memories supplied to the agent satisfy
\[
|\mathcal{R}_t| \leq K_{\max},
\qquad
\mathrm{Len}(\mathcal{R}_t) \leq L_{\max}.
\]
We summarize the overall resource budget as
\[
\mathcal{C}
=
(O_{\max},J_{\max},B_{\max},K_{\max},L_{\max}),
\]
which jointly constrains generation, interaction, memory storage, and memory
usage.

Let $\pi_{\mathrm{mem}}$ denote an online memory policy that determines both
which experiences to retrieve for the current task and how the memory bank is
updated after observing the task outcome.
Through execution feedback, this policy can adapt how accumulated experience
influences future decisions without modifying the backbone.
For a fixed resource budget $\mathcal{C}$, we define its lifelong task
performance as
\[
S(\pi_{\mathrm{mem}};\mathcal{C})
=
\frac{1}{T}
\sum_{t=1}^{T}
\mathbb{E}[r_t].
\]
The objective is therefore
\begin{equation}
\pi_{\mathrm{mem}}^{\star}
=
\arg\max_{\pi_{\mathrm{mem}}\in\Pi(\mathcal{C})}
S(\pi_{\mathrm{mem}};\mathcal{C}),
\label{eq:objective}
\end{equation}
where $\Pi(\mathcal{C})$ denotes the class of online memory policies satisfying
the resource constraints above.

Our focus is on experience-driven adaptation under a stringent joint resource
budget, where generation, interaction, and memory resources are simultaneously
limited. Under such constraints, ineffective memory decisions cannot easily
be compensated for by retrieving additional experience, generating longer
reasoning traces, or interacting for more rounds. Each retrieval and retention
decision therefore becomes more consequential. The central challenge for
resource-efficient self-improvement is not merely to accumulate or retrieve
relevant experience, but to learn which experiences are actually \emph{worth}
retaining and reusing to improve future decisions.
\section{TokenWorth: Feedback-Driven Memory Adaptation}
\label{sec:tokenworth}

\subsection{Overview}
\label{sec:method_overview}

\begin{figure}[t]
    \centering
    \includegraphics[width=\linewidth]{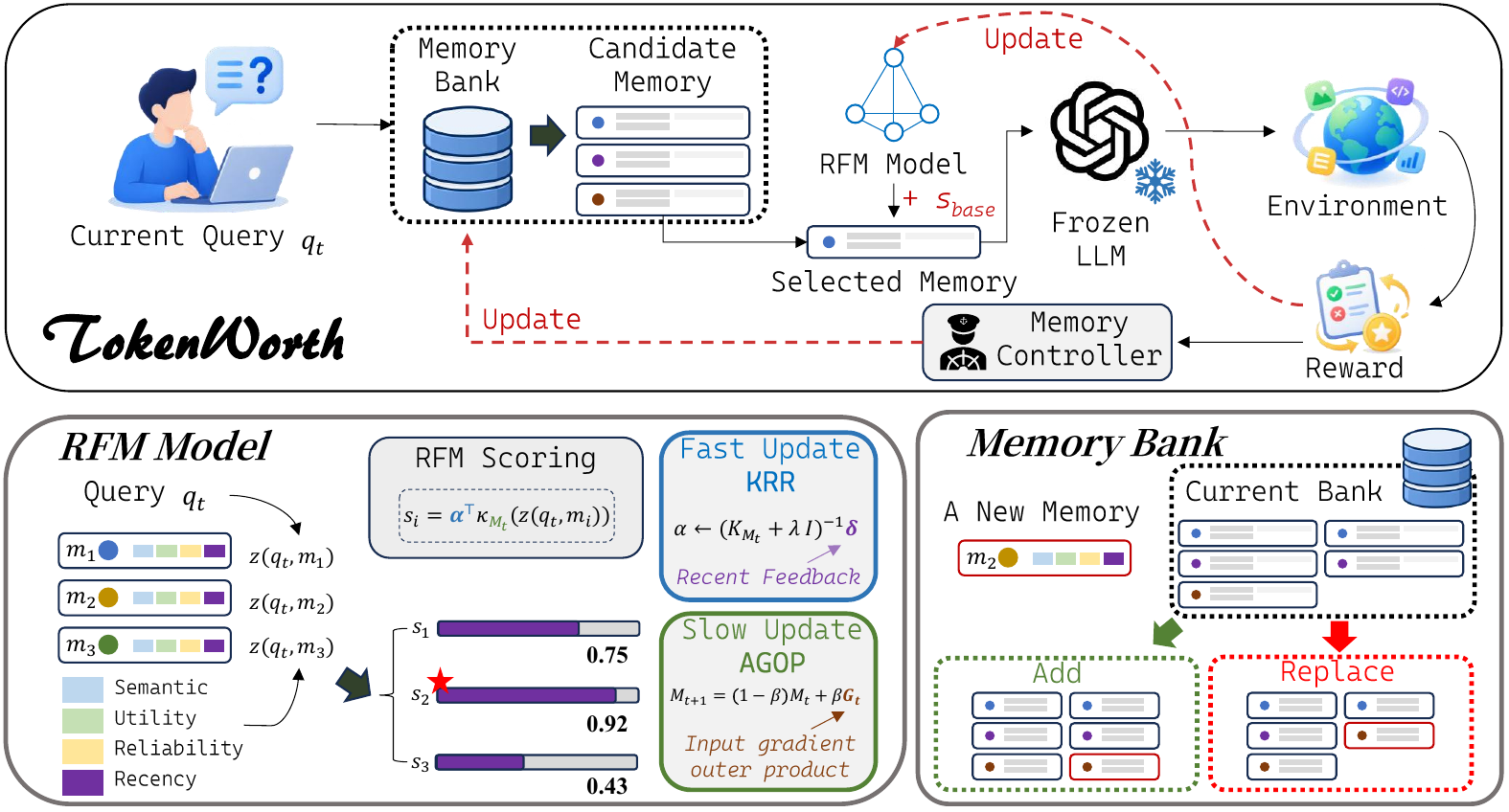}
    \caption{
    Overview of \textbf{TokenWorth}. Online execution feedback updates
    query-dependent memory worth, which guides subsequent experience
    retrieval and bounded memory-bank management.
    }
    \label{fig:method}
    \vspace{-0.3cm}
\end{figure}

Figure~\ref{fig:method} illustrates the feedback-driven adaptation process
of \textbf{TokenWorth}. Given a query $q_t$, the agent first recalls a small
candidate set from its bounded memory bank. TokenWorth estimates the
query-dependent worth of each candidate using a learnable linear base scorer
and an RFM-based residual correction, selecting the highest-valued experience
for task execution. After receiving the environment reward, the worth model
is updated online, and the learned worth further guides memory-bank management.
Through this feedback loop, the agent continually adapts how past experience
influences future decisions without updating its backbone.

TokenWorth consists of three components. First, an interpretable linear
scorer captures stable memory-utility patterns, while a residual Recursive
Feature Machine (RFM) models nonlinear, query-specific corrections
\citep{beaglehole2026xrfm,radhakrishnan2024mechanism}.
Second, fast value updates and slower AGOP-based feature adaptation
balance stability and adaptivity under sparse feedback.
Third, learned worth guides both retrieval and bounded memory management.

\subsection{Learning Query-Dependent Memory Worth}
\label{sec:memory_worth}

Effective experience-driven adaptation requires identifying which past
experiences are worth reusing for future decisions. Unlike a fixed memory-utility score, the usefulness of an
experience depends on the current query
and must be inferred from sparse execution feedback. Each task is executed
only once, feedback is observed only for the memory actually used, and
execution failure may arise from factors beyond the selected memory itself.

To balance stability and adaptivity under such feedback, TokenWorth
decomposes memory worth into two complementary components. A lightweight
linear base scorer captures stable and interpretable patterns across
the task stream, while an RFM learns nonlinear residual structure
not captured by the base scorer. The linear component is updated
conservatively around an informative prior, whereas the RFM adapts
more flexibly to local query--memory interactions.

\paragraph{Candidate recall and representation.}
Each memory $m_i\in\mathcal{B}_t$ contains a compact record of a previously
successful interaction together with lightweight statistics describing its
historical use. Given the current query $q_t$, we first recall a small candidate
set using a lightweight textual relevance function:
\[
\mathcal{C}_t
=
\operatorname{TopK}_{K_c}
\left\{
s_{\mathrm{recall}}(q_t,m)
:
m\in\mathcal{B}_t
\right\},
\]
where $K_c$ controls only the internal candidate pool and is independent of the
number of memories ultimately supplied to the agent.

For each candidate $m_i\in\mathcal{C}_t$, we construct four interpretable
features
\[
\boldsymbol{\phi}_t^i
=
\left[
\phi_{\mathrm{sim}}(q_t,m_i),
\phi_{\mathrm{util}}(m_i),
\phi_{\mathrm{rel}}(m_i),
\phi_{\mathrm{recent}}(m_i)
\right],
\]
capturing lightweight textual similarity, historical utility, reliability, and
recency.

For residual estimation, we additionally construct a richer query--memory
representation
\[
\mathbf{z}_t^i
=
\left[
|\mathbf{e}_t-\mathbf{e}_i|,
\operatorname{cos}(\mathbf{e}_t,\mathbf{e}_i),
u_i,
\operatorname{rel}_t(m_i),
\operatorname{rec}_t(m_i),
\operatorname{freq}_t(m_i)
\right].
\]
This representation combines hashed textual similarity features with the
empirical history of each memory. The text representation is deterministic and
lightweight rather than a learned semantic embedding; detailed definitions are
provided in Appendix~\ref{app:implementation}.

\paragraph{Prior-regularized base scorer.}
The base scorer captures stable global relationships between memory
characteristics and downstream usefulness:
\[
s_{\mathrm{base},t}(q_t,m_i)
=
\mathbf{w}_t^{\top}\boldsymbol{\phi}_t^i.
\]
Rather than learning $\mathbf{w}_t$ from scratch, we initialize it from an
informative prior $\mathbf{w}_0$ and regularize subsequent online updates toward
this prior. Given observed execution feedback
$\mathcal{D}^{\mathrm{base}}_t
=\{(\boldsymbol{\phi}_j^\star,y_j)\}$,
we optimize
\[
\mathcal{L}_{\mathrm{base}}(\mathbf{w})
=
\sum_{(\boldsymbol{\phi}_j^\star,y_j)\in
\mathcal{D}^{\mathrm{base}}_t}
\log\!\left(
1+\exp\!\left(
-y_j\mathbf{w}^{\top}\boldsymbol{\phi}_j^\star
\right)
\right)
+
\lambda_w
\|\mathbf{w}-\mathbf{w}_0\|_2^2.
\]
The prior anchors the scorer when feedback is scarce, while online updates
allow it to capture evolving patterns of memory usefulness.
Conservative periodic updates, smoothing, and bounded weights further
reduce unstable drift; implementation details are provided in
Appendix~\ref{app:implementation}.

\paragraph{Residual worth estimation.}
The linear base scorer captures stable global regularities but cannot model
all nonlinear or query-specific interactions among memory signals.
TokenWorth therefore augments it with an RFM-based residual correction:
\begin{equation}
s_t(q_t,m_i)
=
s_{\mathrm{base},t}(q_t,m_i)
+
\eta_t f_t(\mathbf{z}_t^i),
\label{eq:worth_score}
\end{equation}
where $f_t$ predicts the residual memory worth not explained by the base scorer
and $\eta_t$ controls its contribution. The selected memory is
\[
m_t^\star
=
\arg\max_{m_i\in\mathcal{C}_t}
s_t(q_t,m_i).
\]
The linear scorer captures stable patterns of memory usefulness, while
the RFM models nonlinear, query-specific deviations.

After task execution, the observed reward provides execution-derived
supervision for the selected query--memory pair:
\[
y_t = 2r_t-1 \in \{-1,+1\}.
\]
The base scorer is updated directly from $(\boldsymbol{\phi}_t^\star,y_t)$.
For the residual learner, we first compute the prediction implied by the base
scorer,
\[
\hat y_t^{\mathrm{base}}
=
2\sigma\!\left(
s_{\mathrm{base},t}(q_t,m_t^\star)
\right)-1,
\]
and define the residual target
\[
\delta_t
=
y_t-\hat y_t^{\mathrm{base}}.
\]
The RFM is then updated from $(\mathbf{z}_t^\star,\delta_t)$.
This feedback-driven update allows subsequent memory decisions to incorporate
observed execution outcomes. Importantly, the supervision reflects downstream
task success rather than the counterfactual causal contribution of
$m_t^\star$: TokenWorth does not re-execute the same task with alternative
memories.

\subsection{Two-Timescale RFM Adaptation}
\label{sec:two_timescale_rfm}

Continual adaptation from sparse feedback requires balancing responsiveness
to recent outcomes with stability across the task stream.
Within the residual learner, prediction and feature geometry need not adapt
at the same rate. Recent outcomes provide immediate evidence about local
residual structure, whereas the geometry governing query--memory features
should evolve more conservatively. TokenWorth therefore separates
\emph{fast residual adaptation} from \emph{slow geometry adaptation}.

\paragraph{Fast residual adaptation.}
Given a short feedback buffer
$\mathcal{D}^{\mathrm{fast}}_t
=\{(\mathbf{z}_j,\delta_j)\}_{j=1}^{N_f}$,
we fit a kernel ridge regressor
\begin{equation}
f_t(\mathbf{z})
=
\sum_{j=1}^{N_f}
\alpha_j
K_{\mathbf{M}_t}(\mathbf{z},\mathbf{z}_j),
\qquad
\boldsymbol{\alpha}
=
\left(
\mathbf{K}_{\mathbf{M}_t}
+
\lambda\mathbf{I}
\right)^{-1}
\boldsymbol{\delta},
\label{eq:krr}
\end{equation}
where $\mathbf{M}_t$ is an adaptive diagonal feature metric. The KRR
coefficients are updated frequently, allowing the residual learner to
rapidly capture deviations from the global base scorer.

\paragraph{Slow geometry adaptation.}
A fixed kernel metric would assume that different query--memory features
retain the same relative importance throughout the task stream.
Instead, RFM adapts the feature geometry using the
\emph{Average Gradient Outer Product}
(AGOP)~\citep{radhakrishnan2024mechanism,lai2026golden}.
Given a longer feedback window
$\mathcal{D}^{\mathrm{slow}}_t
=\{(\mathbf{z}_j,\delta_j)\}_{j=1}^{N_s}$,
we compute the diagonal AGOP
\[
\widetilde{\mathbf{G}}_t
=
\operatorname{Diag}
\left(
\frac{1}{N_s}
\sum_{j=1}^{N_s}
\nabla_{\mathbf{z}} f_t(\mathbf{z}_j)
\odot
\nabla_{\mathbf{z}} f_t(\mathbf{z}_j)
\right),
\]
and update the metric as
\begin{equation}
\mathbf{M}_{t+1}
=
\operatorname{NormClip}
\left(
(1-\beta)\mathbf{M}_t
+
\beta\widetilde{\mathbf{G}}_t
\right).
\label{eq:agop_update}
\end{equation}
After each metric update, the KRR predictor is refitted under the updated
geometry.

These adaptation mechanisms play complementary roles in online memory-worth
learning. The prior-regularized linear scorer captures slowly evolving global
regularities, the fast KRR update tracks recent residual structure, and the
slower AGOP update adapts the feature geometry more conservatively.
Together, they aim to balance stability and responsiveness
when learning from limited execution feedback.

\subsection{Worth-Guided Memory Lifecycle}
\label{sec:memory_lifecycle}

Learning which memory to use for the current query addresses only part of
experience-driven adaptation. Under a bounded persistent capacity $B_{\max}$,
the agent must also determine which experiences deserve to remain available
for future tasks. TokenWorth therefore extends the same query-dependent
notion of worth from immediate retrieval to long-term memory management.

Only successful, externally verified interactions are considered as new memory
candidates. For a candidate memory $m$, we estimate its broader worth over a
small set of recent probe queries $\mathcal{P}_t$:
\begin{equation}
V_t(m)
=
\frac{1}{|\mathcal{P}_t|}
\sum_{q\in\mathcal{P}_t}
s_t(q,m).
\label{eq:probe_value}
\end{equation}
While $s_t(q,m)$ measures the query-dependent worth of $m$ for a particular
query, $V_t(m)$ aggregates this signal over recent queries to estimate whether
the memory deserves persistent capacity under the current task distribution.

TokenWorth follows a coverage--quality principle for memory admission.
When capacity remains available, successful experiences are admitted
to build initial coverage. Once the bank is full, a candidate is admitted
if it is sufficiently novel, or if its probe-averaged worth exceeds that
of the best existing memory and satisfies a positive-score ratio criterion.
Candidates failing this admission gate may still replace a low-utility
nearest neighbor when they provide a sufficiently large worth improvement.
Detailed admission and replacement criteria are provided in
Appendix~\ref{app:implementation}.

For candidates passing the admission gate, TokenWorth enforces the capacity
bound by evicting the memory with the lowest probe-averaged worth from the
enlarged bank:
\[
m_{\mathrm{evict}}
=
\arg\min_{m\in
\mathcal{B}_t\cup\{m_{\mathrm{new}}\}}
V_t(m).
\]

TokenWorth thus applies a unified notion of memory worth at two complementary
timescales: query-dependent worth determines which experience should be used
\emph{now}, while probe-averaged worth determines which experience should be
retained for \emph{later}. This feedback-driven memory lifecycle continually
reallocates limited capacity toward experiences estimated to be useful for
future tasks, allowing the agent to adapt its experience-reuse policy over
time. All adaptation occurs within the lightweight memory controller;
TokenWorth requires no backbone updates, task replay, counterfactual
execution, or additional language-model calls.
\section{Experiments}

\subsection{Experimental Setup}

\noindent \textbf{Benchmarks.}
We conduct our main evaluation on three environments from
LifelongAgentBench~\citep{zheng2025lifelongagentbench}: database interaction
(DB), operating-system interaction (OS), and knowledge-graph interaction (KG).
These environments cover heterogeneous forms of interactive decision making
under an online lifelong protocol, where agents can continually adapt their
experience-reuse policies based on preceding interactions.
In all environments, the memory bank is initialized empty, each task is
executed once, and memory decisions depend only on experience observed
from previous tasks.

We additionally evaluate generalization beyond LifelongAgentBench using online
lifelong variants of BigCodeBench~\citep{zhuo2025bigcodebench} and
ALFWorld~\citep{shridhar2020alfworld}. These results are reported in
Appendix~\ref{app:generalization}.

\noindent \textbf{Backbone and deployment.}
We evaluate TokenWorth under two complementary deployment settings.
In the \emph{local} setting, all methods use
Qwen2.5-14B-Instruct~\citep{qwen2.5} as the frozen backbone, where generation
and interaction are constrained by computation and latency.
In the \emph{API-based} setting, all methods use
DeepSeek-V4-Flash~\citep{xu2026deepseek} through an external API, where model
generation and repeated interaction directly incur inference cost.
The backbone remains frozen throughout evaluation, so adaptation occurs
solely through external memory mechanisms.

\noindent \textbf{Resource-constrained protocol.}
Following Sec.~\ref{sec:problem_formulation}, we evaluate all methods under a
stringent joint resource budget that constrains generation, interaction,
persistent memory capacity, and memory injection.
Within each environment and deployment setting, all methods process the same
task stream under the same resource budget. All language-model generation
incurred by a method, including auxiliary calls for memory construction,
reflection, or skill extraction, is counted toward the same per-task generation
budget $O_{\max}$.
This controlled protocol allows us to examine the effectiveness of
experience-driven adaptation when both task execution and experience reuse
are severely constrained. Across DB, OS, and KG, memory capacity is 64 and
at most one memory is retrieved per task. The respective output-token budgets
are 128, 512, and 128, with interaction limits of 3, 5, and 15 rounds.
These results characterize effectiveness under this fixed stringent budget;
we do not claim robustness across a range of budget levels.

\noindent \textbf{Baselines.}
We compare TokenWorth with a no-memory agent (Standard), vanilla
retrieval-augmented generation (RAG)~\citep{lewis2020retrieval}, and
representative lifelong-memory methods, including
MemRL~\citep{zhang2026memrl},
Mem0~\citep{chhikara2025mem0},
MemP~\citep{fang2026memp},
MemSkill~\citep{zhang2026memskill},
Reflexion~\citep{shinn2023reflexion},
EvoSC~\citep{yu2026self}, and
EvoSkill~\citep{alzubi2026evoskill}.
All methods share the same frozen backbone, task stream, and resource budget
within each setting.

\noindent \textbf{Metrics.}
We evaluate task effectiveness using task success rate
($\mathrm{SR}\uparrow$).
To measure generation efficiency, we report the average number of output
tokens per task (OutputTok $\downarrow$), including all language-model
generation incurred during task execution and method-specific auxiliary
operations such as memory construction, reflection, or skill extraction.
We further report the relative change in output-token usage with respect to
the no-memory Standard agent:
\[
\Delta\mathrm{Tok}
=
\frac{
\mathrm{OutputTok}_{\mathrm{method}}
-
\mathrm{OutputTok}_{\mathrm{Standard}}
}{
\mathrm{OutputTok}_{\mathrm{Standard}}
}
\times 100\%,
\]
where a negative value indicates lower generation consumption than the
no-memory baseline.

\subsection{Results and Analysis}

\begin{table*}[t]
\centering
\caption{
Task success rate (\%) on LifelongAgentBench under stringent joint resource
constraints. Best results are shown in \textbf{bold} and second-best results
are \underline{underlined}.
}
\label{tab:main_results}

\small
\setlength{\tabcolsep}{9pt}
\renewcommand{\arraystretch}{1.10}

\begin{tabular}{@{}lcccccc@{}}
\toprule
& \multicolumn{3}{c}{\textit{Local: Qwen2.5-14B-Instruct}}
& \multicolumn{3}{c}{\textit{API: DeepSeek-V4-Flash}} \\
\cmidrule(lr){2-4}
\cmidrule(lr){5-7}

Method
& \textsc{DB}
& \textsc{OS}
& \textsc{KG}
& \textsc{DB}
& \textsc{OS}
& \textsc{KG} \\
\midrule

Standard
& 25.2
& 50.2
& \underline{30.7}
& 26.2
& 67.2
& 12.7 \\

RAG
& \underline{33.0}
& \underline{54.6}
& 11.3
& 29.4
& 66.6
& 31.3 \\

MemRL
& 28.0
& 52.2
& 4.7
& 30.4
& 66.2
& \underline{32.0} \\

Mem0
& 21.2
& 50.6
& \underline{30.7}
& \underline{32.2}
& 68.8
& 15.3 \\

MemP
& 23.4
& 46.8
& 27.3
& 28.0
& 69.8
& 21.3 \\

MemSkill
& 24.4
& 50.8
& 26.0
& 27.4
& 66.0
& 20.7 \\

Reflexion
& 22.0
& 48.8
& 30.0
& 30.6
& 67.2
& 14.7 \\

EvoSC
& 19.2
& 50.2
& 27.3
& 30.6
& \underline{70.2}
& 18.7 \\

EvoSkill
& 24.6
& 49.6
& 30.0
& 25.6
& 68.4
& 14.7 \\

\textbf{TokenWorth}
& \textbf{65.0}
& \textbf{59.4}
& \textbf{36.7}
& \textbf{62.2}
& \textbf{73.2}
& \textbf{36.7} \\

\bottomrule
\end{tabular}
\end{table*}

\noindent \textbf{RQ1: Can TokenWorth support effective experience-driven adaptation under stringent resource constraints?}
Table~\ref{tab:main_results} reports the main results on LifelongAgentBench.
TokenWorth achieves the highest task success rate in all six
environment--deployment combinations, demonstrating the effectiveness of
its feedback-driven memory adaptation framework across both local and
API-based backbones.

Under the local deployment, the improvement is particularly pronounced on DB,
where TokenWorth reaches $65.0\%$, outperforming the strongest baseline RAG
($33.0\%$) by $32.0$ percentage points. It also improves the strongest baseline
by $4.8$ points on OS and $6.0$ points on KG.
The same pattern holds under the API-based deployment: TokenWorth improves the
best competing method by $30.0$ points on DB, $3.0$ points on OS, and $4.7$
points on KG.
Notably, several memory-based methods provide limited gains over the no-memory
Standard agent and can even substantially underperform it, particularly on KG.
These results suggest that accumulating and reusing experience alone does not
guarantee effective adaptation. Learning which experiences are worth reusing
is particularly important when generation, interaction, and memory resources
are jointly constrained.

\begin{figure*}[t]
    \centering

    \begin{minipage}[t]{0.60\textwidth}
        \centering
        \includegraphics[width=\linewidth]{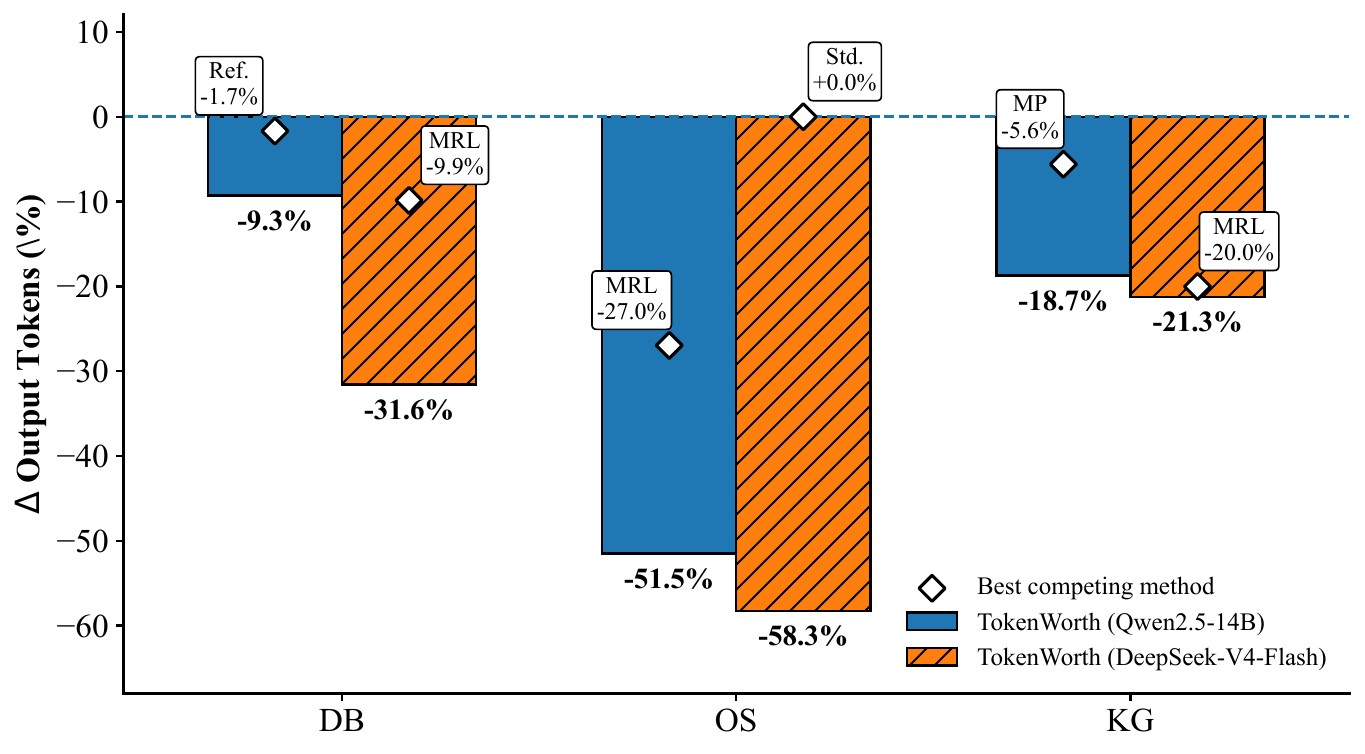}
        \vspace{-2mm}
        \centerline{\small (a) Output-token efficiency}
    \end{minipage}
    \hfill
    \begin{minipage}[t]{0.36\textwidth}
        \centering
        \includegraphics[width=\linewidth]{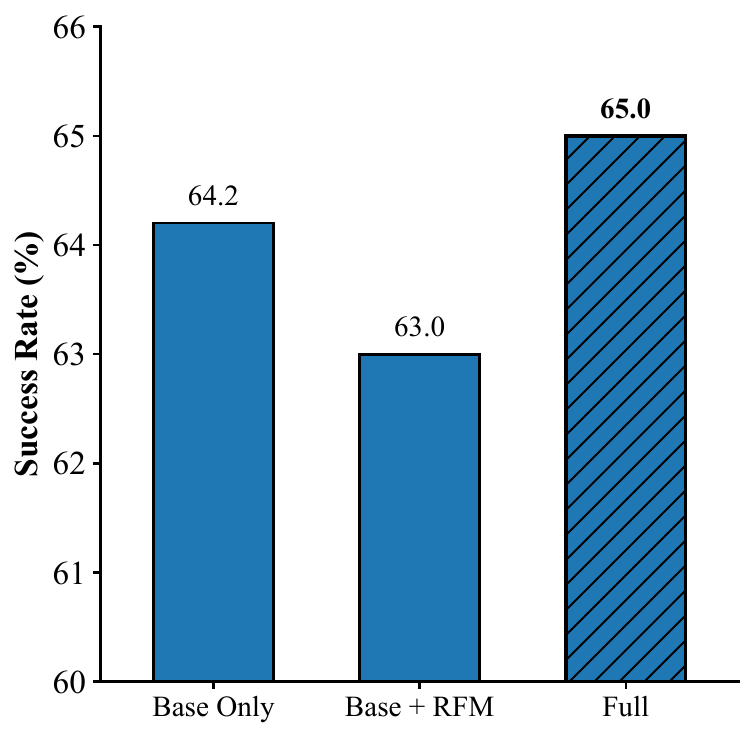}
        \vspace{-2mm}
        \centerline{\small (b) Core component ablation}
    \end{minipage}

    \caption{
    Efficiency and component analysis of TokenWorth.
    \textbf{(a)} Output-token usage relative to the no-memory Standard agent;
    diamonds indicate the most token-efficient competitor in each setting.
    \textbf{(b)} Core ablation on DB with the local backbone. The truncated
    vertical axis should be interpreted alongside the reported values.
    }
    \label{fig:efficiency_ablation}
\end{figure*}

\noindent \textbf{RQ2: Can effective experience reuse improve generation efficiency?}
Figure~\ref{fig:efficiency_ablation}(a) compares TokenWorth's output-token
usage across the six LifelongAgentBench environment--deployment combinations.
We report the relative change with respect to the no-memory Standard agent,
while the diamonds indicate the most token-efficient competing method.
TokenWorth achieves the lowest output-token usage in all six settings.

Under the local Qwen2.5-14B-Instruct deployment, TokenWorth reduces output
generation by $9.3\%$, $51.5\%$, and $18.7\%$ on DB, OS, and KG,
respectively. Under DeepSeek-V4-Flash, the corresponding reductions are
$31.6\%$, $58.3\%$, and $21.3\%$. The reduction is particularly substantial
on OS, where TokenWorth uses less than half the output tokens of the Standard
agent under both deployments.

These savings accompany the highest success rate in all six settings
(Table~\ref{tab:main_results}). TokenWorth also uses fewer output tokens than
the most generation-efficient baseline in each setting, with clear margins
on DB and OS. Although the margin over MemRL on API-based KG is smaller,
TokenWorth still improves both success and token usage.
These results suggest that learning to reuse higher-value experience helps
the frozen backbone solve more tasks with less generation, supporting
resource-efficient experience-driven adaptation.

\noindent \textbf{RQ3: How do the adaptation components affect memory-worth learning?}
Figure~\ref{fig:efficiency_ablation}(b) reports a core ablation on
DB with Qwen2.5-14B-Instruct. We compare the base scorer alone
(\textit{Base Only}), the base scorer augmented with an RFM residual
learner using a fixed feature metric (\textit{Base + RFM}), and the
full framework with AGOP-based feature adaptation (\textit{Full}).

The base scorer achieves $64.2\%$, demonstrating the effectiveness
of stable, interpretable memory signals. Adding the RFM residual
learner with a fixed feature metric reduces performance to
$63.0\%$, indicating that nonlinear residual modeling alone
does not necessarily improve memory-worth estimation.
With AGOP-based feature adaptation, the full framework reaches
$65.0\%$, recovering this degradation and modestly outperforming
the base-only variant. These results suggest that adaptive feature
geometry improves residual corrections under sparse execution
feedback, highlighting the importance of balancing stability
and adaptivity in online memory-worth learning.
\section{Conclusion}

We study experience-driven adaptation in lifelong agents under joint
constraints on generation, interaction, and memory. We propose
\textbf{TokenWorth}, a lightweight feedback-driven memory adaptation
framework that learns query-dependent memory worth from online execution
feedback and uses it for retrieval and bounded memory management.
Across three LifelongAgentBench environments and local and API-based
backbones, TokenWorth achieves the highest task success rates among
evaluated methods while using the fewest output tokens. Additional
experiments show competitive performance on coding and embodied-agent
tasks. These results suggest that resource-efficient self-improvement
depends not on accumulating more experience, but on learning what is
worth remembering. Since feedback is observed only for selected memories,
their causal contributions remain unidentified. Developing budget-efficient
credit assignment remains an important direction for future work. 

\begin{ack}
This work was supported by the National Key R\&D Program of China under
Grant No. 2024YFE0202800 and the National Natural Science Foundation of China
under Grant Nos. 62522605 and 62376118.
\end{ack}

\bibliographystyle{plainnat}
\bibliography{references}

@inproceedings{lai2025order,
  title={Order-robust class incremental learning: Graph-driven dynamic similarity grouping},
  author={Lai, Guannan and Li, Yujie and Wang, Xiangkun and Zhang, Junbo and Li, Tianrui and Yang, Xin},
  booktitle={2025 IEEE/CVF Conference on Computer Vision and Pattern Recognition (CVPR)},
  pages={4894--4904},
  year={2025},
  organization={IEEE}
}

@inproceedings{lai2026lie,
  title={The Lie of the Average: How Class Incremental Learning Evaluation Deceives You?},
  author={Lai, Guannan and Zhou, Da-Wei and Yang, Xin and Ye, Han-Jia},
  booktitle={International Conference on Learning Representations},
  volume={2026},
  pages={110530--110562},
  year={2026}
}

@article{zheng2026lifelong,
  title={Lifelong learning of large language model based agents: A roadmap},
  author={Zheng, Junhao and Shi, Chengming and Cai, Xidi and Li, Qiuke and Zhang, Duzhen and Li, Chenxing and Yu, Dong and Ma, Qianli},
  journal={TPAMI},
  year={2026},
  publisher={IEEE}
}

@article{fang2025comprehensive,
  title={A comprehensive survey of self-evolving ai agents: A new paradigm bridging foundation models and lifelong agentic systems},
  author={Fang, Jinyuan and Peng, Yanwen and Zhang, Xi and Wang, Yingxu and Yi, Xinhao and Zhang, Guibin and Xu, Yi and Wu, Bin and Liu, Siwei and Li, Zihao and others},
  journal={arXiv preprint arXiv:2508.07407},
  year={2025}
}

@article{zheng2025lifelongagentbench,
  title={Lifelongagentbench: Evaluating llm agents as lifelong learners},
  author={Zheng, Junhao and Cai, Xidi and Li, Qiuke and Zhang, Duzhen and Li, ZhongZhi and Zhang, Yingying and Song, Le and Ma, Qianli},
  journal={arXiv preprint arXiv:2505.11942},
  year={2025}
}

@inproceedings{luo2025agrail,
  title={Agrail: A lifelong agent guardrail with effective and adaptive safety detection},
  author={Luo, Weidi and Dai, Shenghong and Liu, Xiaogeng and Banerjee, Suman and Sun, Huan and Chen, Muhao and Xiao, Chaowei},
  booktitle={ACL},
  year={2025}
}

@inproceedings{cheng2024sociodojo,
  title={Sociodojo: Building lifelong analytical agents with real-world text and time series},
  author={Cheng, Junyan and Chin, Peter},
  booktitle={ICLR},
  year={2024}
}

@inproceedings{jiang2026xskill,
  title={Xskill: Continual learning from experience and skills in multimodal agents},
  author={Jiang, Guanyu and Su, Zhaochen and Qu, Xiaoye and Fung, Yi R},
  booktitle={ICML},
  year={2026}
}

@article{shu2026agentcl,
  title={AgentCL: Toward Rigorous Evaluation of Continual Learning in Language Agents},
  author={Shu, Yiheng and Guti{\'e}rrez, Bernal Jim{\'e}nez and Jonnalagedda, Saisri Padmaja and Yao, Yuguang and Sun, Huan and Su, Yu},
  journal={arXiv preprint arXiv:2606.02461},
  year={2026}
}

@inproceedings{beaglehole2026xrfm,
  title={xRFM: Accurate, scalable, and interpretable feature learning models for tabular data},
  author={Beaglehole, Daniel and Holzm{\"u}ller, David and Radhakrishnan, Adityanarayanan and Belkin, Misha},
  booktitle={International Conference on Learning Representations},
  volume={2026},
  pages={146422--146454},
  year={2026}
}

@inproceedings{zhong2024memorybank,
  title={Memorybank: Enhancing large language models with long-term memory},
  author={Zhong, Wanjun and Guo, Lianghong and Gao, Qiqi and Ye, He and Wang, Yanlin},
  booktitle= {AAAI},
  year={2024}
}

@article{shinn2023reflexion,
  title={Reflexion: Language agents with verbal reinforcement learning},
  author={Shinn, Noah and Cassano, Federico and Gopinath, Ashwin and Narasimhan, Karthik and Yao, Shunyu},
  journal={NeurIPS},
  year={2023}
}

@inproceedings{zhao2024expel,
  title={Expel: Llm agents are experiential learners},
  author={Zhao, Andrew and Huang, Daniel and Xu, Quentin and Lin, Matthieu and Liu, Yong-Jin and Huang, Gao},
  booktitle={AAAI},
  year={2024}
}

@article{wangvoyager,
  title={Voyager: An Open-Ended Embodied Agent with Large Language Models},
  author={Wang, Guanzhi and Xie, Yuqi and Jiang, Yunfan and Mandlekar, Ajay and Xiao, Chaowei and Zhu, Yuke and Fan, Linxi and Anandkumar, Anima},
  journal={TMLR},
  year = {2023}
}

@article{zhong2026skilllearnbench,
  title={SkillLearnBench: Benchmarking Continual Learning Methods for Agent Skill Generation on Real-World Tasks},
  author={Zhong, Shanshan and Lu, Yi and Ning, Jingjie and Wan, Yibing and Feng, Lihan and Ao, Yuyi and Ribeiro, Leonardo FR and Dreyer, Markus and Ammirati, Sean and Xiong, Chenyan},
  journal={arXiv preprint arXiv:2604.20087},
  year={2026}
}

@article{lewis2020retrieval,
  title={Retrieval-augmented generation for knowledge-intensive nlp tasks},
  author={Lewis, Patrick and Perez, Ethan and Piktus, Aleksandra and Petroni, Fabio and Karpukhin, Vladimir and Goyal, Naman and K{\"u}ttler, Heinrich and Lewis, Mike and Yih, Wen-tau and Rockt{\"a}schel, Tim and others},
  journal={NeurIPS},
  year={2020}
}

@article{chhikara2025mem0,
  title={Mem0: Building production-ready ai agents with scalable long-term memory},
  author={Chhikara, Prateek and Khant, Dev and Aryan, Saket and Singh, Taranjeet and Yadav, Deshraj},
  journal={arXiv preprint arXiv:2504.19413},
  year={2025}
}

@article{zhang2026memrl,
  title={Memrl: Self-evolving agents via runtime reinforcement learning on episodic memory},
  author={Zhang, Shengtao and Wang, Jiaqian and Zhou, Ruiwen and Liao, Junwei and Feng, Yuchen and Li, Zhuo and Zheng, Yujie and Zhang, Weinan and Wen, Ying and Li, Zhiyu and others},
  journal={arXiv preprint arXiv:2601.03192},
  year={2026}
}

@article{zhang2026memskill,
  title={MemSkill: Learning and Evolving Memory Skills for Self-Evolving Agents},
  author={Zhang, Haozhen and Long, Quanyu and Bao, Jianzhu and Feng, Tao and Zhang, Weizhi and Yue, Haodong and Wang, Wenya},
  journal={arXiv preprint arXiv:2602.02474},
  year={2026}
}

@inproceedings{kwon2023efficient,
  title={Efficient memory management for large language model serving with pagedattention},
  author={Kwon, Woosuk and Li, Zhuohan and Zhuang, Siyuan and Sheng, Ying and Zheng, Lianmin and Yu, Cody Hao and Gonzalez, Joseph and Zhang, Hao and Stoica, Ion},
  booktitle={Proceedings of the 29th symposium on operating systems principles},
  pages={611--626},
  year={2023}
}

@article{radhakrishnan2024mechanism,
  title={Mechanism for feature learning in neural networks and backpropagation-free machine learning models},
  author={Radhakrishnan, Adityanarayanan and Beaglehole, Daniel and Pandit, Parthe and Belkin, Mikhail},
  journal={Science},
  volume={383},
  number={6690},
  pages={1461--1467},
  year={2024},
  publisher={American Association for the Advancement of Science}
}

@article{gupta2025ai,
  title={Ai agents collaboration under resource constraints: Practical implementations},
  author={Gupta, Shubham},
  journal={International journal of artificial intelligence research and development},
  volume={3},
  number={1},
  pages={51--63},
  year={2025}
}

@article{yuan2026large,
  title={Large Language Models over Networks: Collaborative Intelligence under Resource Constraints},
  author={Yuan, Liangqi and Fang, Wenzhi and Wang, Shiqiang and Poor, H Vincent and Brinton, Christopher G},
  journal={arXiv preprint arXiv:2605.08626},
  year={2026}
}

@article{yang2026autoskill,
  title={Autoskill: Experience-driven lifelong learning via skill self-evolution},
  author={Yang, Yutao and Li, Junsong and Pan, Qianjun and Zhan, Bihao and Cai, Yuxuan and Du, Lin and Zhou, Jie and Chen, Kai and Chen, Qin and Li, Xin and others},
  journal={arXiv preprint arXiv:2603.01145},
  year={2026}
}

@article{hajimiri2026online,
  title={Are Online Skill and Memory Modules Always Worth Their Tokens? A Budget-Constrained Study of Web Agents},
  author={Hajimiri, Sina and Aminbeidokhti, Masih and Dolz, Jose and Ayed, Ismail Ben and Laradji, Issam H and Gella, Spandana and Gontier, Nicolas},
  journal={arXiv preprint arXiv:2606.15017},
  year={2026}
}

@article{wu2026forget,
  title={Forget to Improve: On-Device LLM-Agent Continual Learning via Budget-Curated Memory},
  author={Wu, Beining and Ding, Zihao and Huang, Jun and Zhao, Yanxiao},
  journal={arXiv preprint arXiv:2606.25115},
  year={2026}
}

@inproceedings{fang2026memp,
  title={Memp: Exploring agent procedural memory},
  author={Fang, Runnan and Liang, Yuan and Wang, Xiaobin and Wu, Jialong and Qiao, Shuofei and Xie, Pengjun and Huang, Fei and Chen, Huajun and Zhang, Ningyu},
  booktitle={Findings of the Association for Computational Linguistics: ACL 2026},
  pages={17490--17502},
  year={2026}
}

@article{yu2026self,
  title={Self-consolidation for self-evolving agents},
  author={Yu, Hongzhuo and Zhu, Fei and Xie, Guo-Sen and Shao, Ling},
  journal={arXiv preprint arXiv:2602.01966},
  year={2026}
}

@article{alzubi2026evoskill,
  title={Evoskill: Automated skill discovery for multi-agent systems},
  author={Alzubi, Salaheddin and Provenzano, Noah and Bingham, Jaydon and Chen, Weiyuan and Vu, Tu},
  journal={arXiv preprint arXiv:2603.02766},
  year={2026}
}

@article{zhang2026learning,
  title={Learning query-aware budget-tier routing for runtime agent memory},
  author={Zhang, Haozhen and Yue, Haodong and Feng, Tao and Long, Quanyu and Bao, Jianzhu and Jin, Bowen and Zhang, Weizhi and Li, Xiao and You, Jiaxuan and Qin, Chengwei and others},
  journal={arXiv preprint arXiv:2602.06025},
  year={2026}
}

@article{chen2026learning,
  title={Learning What to Remember: A Cognitively Grounded Multi-Factor Value Model for Agentic Memory},
  author={Chen, Zhibao and Cheng, Qian},
  journal={arXiv preprint arXiv:2606.12945},
  year={2026}
}

@article{sun2026scaling,
  title={Scaling DPPs for RAG: Density Meets Diversity},
  author={Sun, Xun and Xie, Baiheng and Huang, Li and Gao, Qiang},
  journal={arXiv preprint arXiv:2604.03240},
  year={2026}
}

@article{sun2026m,
  title={M-RAG: Semantic Key-Value Indexing for Retrieval-Augmented Generation},
  author={Sun, Xu and Xu, Tongkai and Xie, Baiheng and Huang, Li and Gao, Qiang and Zhang, Kunpeng},
  journal={arXiv e-prints},
  pages={arXiv--2603},
  year={2026}
}

@inproceedings{zhuo2025bigcodebench,
  title={Bigcodebench: Benchmarking code generation with diverse function calls and complex instructions},
  author={Zhuo, Terry Yue and Vu, Minh Chien and Chim, Jenny and Hu, Han and Yu, Wenhao and Widyasari, Ratnadira and Yusuf, Imam Nur Bani and Zhan, Haolan and He, Junda and Paul, Indraneil and others},
  booktitle={International Conference on Learning Representations},
  volume={2025},
  pages={66602--66656},
  year={2025}
}

@article{shridhar2020alfworld,
  title={Alfworld: Aligning text and embodied environments for interactive learning},
  author={Shridhar, Mohit and Yuan, Xingdi and C{\^o}t{\'e}, Marc-Alexandre and Bisk, Yonatan and Trischler, Adam and Hausknecht, Matthew},
  journal={arXiv preprint arXiv:2010.03768},
  year={2020}
}

@article{xu2026deepseek,
  title={Deepseek-v4: Towards highly efficient million-token context intelligence},
  author={Xu, Anyi and Lin, Bangcai and Xue, Bing and Wang, Bingxuan and Xu, Bingzheng and Wu, Bochao and Zhang, Bowei and Lin, Chaofan and Dong, Chen and Ling, Chenchen and others},
  journal={arXiv preprint arXiv:2606.19348},
  year={2026}
}

@misc{qwen2.5,
    title = {Qwen2.5: A Party of Foundation Models},
    url = {https://qwenlm.github.io/blog/qwen2.5/},
    author = {{Qwen Team}},
    month = {September},
    year = {2024}
}

@inproceedings{lai2026golden,
  title     = {The Golden Subspace: Where Efficiency Meets Generalization in Continual Test-Time Adaptation},
  author    = {Lai, Guannan and Zhou, Da-Wei and Li, Zhenguo and Ye, Han-Jia},
  booktitle = {CVPR},
  year      = {2026}
}

\appendix
\clearpage
\appendix
\setcounter{figure}{0}
\setcounter{table}{0}
\renewcommand{\thefigure}{A\arabic{figure}}
\renewcommand{\thetable}{A\arabic{table}}

\section{Additional Experimental Details}
\label{app:experimental_details}

\subsection{Benchmarks}
Our main experiments use three environments from
LifelongAgentBench~\citep{zheng2025lifelongagentbench}: database interaction
(DB), operating-system interaction (OS), and knowledge-graph interaction (KG).
Tasks are processed sequentially, and only experience from preceding tasks may
be reused. DB requires structured database operations and answers grounded in
retrieved observations. OS requires multi-step system management through Bash
commands. KG follows an AgentBench-style protocol in which the agent interacts
with a knowledge graph through structured tool calls, while SPARQL queries
are executed by the backend. The environments thus cover heterogeneous
forms of interactive decision making under a shared online
lifelong-learning protocol.

We additionally evaluate online lifelong variants of
BigCodeBench~\citep{zhuo2025bigcodebench} and
ALFWorld~\citep{shridhar2020alfworld}; their results are reported in
Appendix~\ref{app:generalization}.

Beyond language agents, class-incremental learning has examined
the impact of task ordering on continual adaptation
\citep{lai2025order} and the variability of performance across
evaluation sequences \citep{lai2026lie}.
These studies highlight the importance of evaluation protocols
in continual learning, while our experiments focus on online
memory adaptation under fixed resource budgets.

\subsection{Baselines}
We compare TokenWorth with a no-memory agent (Standard),
retrieval-augmented generation (RAG), and representative memory- and
skill-learning methods.

\paragraph{Standard.}
The frozen backbone solves every task independently, without persistent
cross-task memory.
\paragraph{RAG.}
RAG~\citep{lewis2020retrieval,sun2026m,sun2026scaling} stores successful
experiences and retrieves them by lightweight textual relevance, without
online value learning.
\paragraph{MemRL.}
MemRL~\citep{zhang2026memrl} uses execution feedback to learn utility
signals for episodic memories, serving as a value-aware baseline.
\paragraph{Mem0.}
Mem0~\citep{chhikara2025mem0} supports memory extraction, storage,
retrieval, and update in long-running agents.
\paragraph{MemP.}
MemP~\citep{fang2026memp} retains procedural task experience for later reuse.
\paragraph{MemSkill.}
MemSkill~\citep{zhang2026memskill} converts experience into reusable memory
skills and evolves them online.
\paragraph{Reflexion.}
Reflexion~\citep{shinn2023reflexion} stores verbal reflections derived from
task feedback for reuse in subsequent tasks.
\paragraph{EvoSC.}
EvoSC~\citep{yu2026self} consolidates and evolves agent experience.
\paragraph{EvoSkill.}
EvoSkill~\citep{alzubi2026evoskill} extracts reusable skills from interaction
trajectories and updates its skill repository.

Within each environment and deployment setting, methods share the same task
sequence, backbone, environment interface, and resource budget. We preserve
baseline-specific auxiliary operations such as reflection, memory construction,
and skill extraction; their generated output tokens are included in both the
per-task budget and the reported OutputTok. Baseline implementations are
adapted to satisfy the common resource limits.

\subsection{Backbones and Deployment Settings}
\paragraph{Local deployment.}
All methods use the frozen Qwen2.5-14B-Instruct~\citep{qwen2.5}
backbone on NVIDIA A800 GPUs.
\paragraph{API-based deployment.}
All methods use the frozen DeepSeek-V4-Flash~\citep{xu2026deepseek}
backbone through an external inference API.

In both deployments, the memory bank starts empty, task order is fixed within
an environment, and no backbone parameters are updated. TokenWorth uses
neither offline pretraining nor task replay or counterfactual re-execution.

\subsection{Evaluation Protocol}
We measure task effectiveness with task success rate ($\mathrm{SR}\uparrow$).
A task succeeds only if the environment evaluator marks its final answer or
state as correct. Generation efficiency is measured by average output tokens
per task (OutputTok $\downarrow$), including tokens from task execution and
all method-specific auxiliary language-model calls. Relative to Standard,
\begin{equation*}
\Delta\mathrm{Tok}=\frac{\mathrm{OutputTok}_{\mathrm{method}}-
\mathrm{OutputTok}_{\mathrm{Standard}}}
{\mathrm{OutputTok}_{\mathrm{Standard}}}\times100\%.
\end{equation*}
Negative values indicate fewer generated tokens. OutputTok does not include
input tokens or controller-side computation and should not be interpreted as
total inference cost.

\section{Implementation Details of TokenWorth}
\label{app:implementation}

\subsection{Common Online Implementation}

\paragraph{Online feedback protocol.}
At the beginning of task $t$, retrieval uses the bank $\mathcal{B}_t$ and
feedback available through task $t-1$. Task $t$ is executed once. Following
its completion, the observed reward updates the worth model and the selected
memory's statistics, and memory admission or eviction constructs
$\mathcal{B}_{t+1}$. These updates may affect task $t+1$ and later tasks;
there is no same-task counterfactual execution or backbone fine-tuning.

\paragraph{Lightweight text representation.}
We construct deterministic $512$-dimensional text vectors by token hashing
and character $3$--$5$-gram hashing, followed by $\ell_2$ normalization.
This representation requires no learned encoder and captures lightweight
textual similarity rather than learned semantic embeddings. For a query $q$
and memory $m$, the residual input is
\begin{equation*}
\mathbf{z}(q,m)=\bigl[|\mathbf{e}_q-\mathbf{e}_m|,
\operatorname{cos}(\mathbf{e}_q,\mathbf{e}_m),u_m,
\operatorname{rel}(m),\operatorname{rec}(m),\operatorname{freq}(m)\bigr].
\end{equation*}
Here $u_m$ is historical utility. With $h_m$ and $d_m$ denoting counts of
successful and unsuccessful executions, respectively, \emph{conditioned on
selecting $m$}, the other statistics are
\begin{align*}
\operatorname{rel}(m)&=\frac{h_m+1}{h_m+d_m+2},\\
\operatorname{rec}(m)&=\exp(-\kappa\Delta t_m),\\
\operatorname{freq}(m)&=\frac{c_m}{\max(1,t-\tau_m+1)}.
\end{align*}
Here $\Delta t_m$ is the number of steps since last access, $c_m$ is the
access count, and $\tau_m$ is the creation step. We set $\kappa=0.02$.
The success and failure counts are execution-derived statistics, not causal
estimates of a memory's effect.

\paragraph{Candidate recall and base scorer.}
The bank is screened by lightweight textual relevance to form
\begin{equation*}
\mathcal{C}_t=\operatorname{TopK}_{K_c}
\{s_{\mathrm{recall}}(q_t,m):m\in\mathcal{B}_t\}.
\end{equation*}
For each candidate, define
$\boldsymbol{\phi}(q,m)=[\phi_{\mathrm{sim}},\phi_{\mathrm{util}},
\phi_{\mathrm{rel}},\phi_{\mathrm{recent}}]$ and
$s_{\mathrm{base},t}(q,m)=\mathbf{w}_t^\top\boldsymbol{\phi}(q,m)$.
We initialize $\mathbf{w}_0=[1.0,0.10,0.10,0.02]$ and fit
\begin{equation*}
\mathcal{L}_{\mathrm{base}}(\mathbf{w})=
\sum_{(\boldsymbol{\phi}_j,y_j)\in\mathcal{D}^{\mathrm{base}}_t}
\log\!\left(1+\exp(-y_j\mathbf{w}^\top\boldsymbol{\phi}_j)\right)
+\lambda_w\|\mathbf{w}-\mathbf{w}_0\|_2^2.
\end{equation*}
The FIFO training window contains at most $256$ selected query--memory pairs,
with $y_j\in\{-1,+1\}$. We use $\lambda_w=0.1$ and refit after every
$50$ new feedback examples using deterministic Newton optimization with
backtracking line search. Updated weights are smoothed with coefficient
$\beta_w=0.5$ and clipped coordinate-wise between
$[0,0,0,0]$ and $[2.0,1.0,1.0,1.0]$.

\paragraph{Residual worth estimation.}
The combined score and selected memory are
\begin{equation*}
s_t(q,m)=s_{\mathrm{base},t}(q,m)+\eta_t f_t(\mathbf{z}(q,m)),
\qquad m_t^\star=\arg\max_{m\in\mathcal{C}_t}s_t(q_t,m).
\end{equation*}
To avoid abrupt residual corrections during early adaptation, we linearly
increase the residual coefficient over the first three KRR fits:
\begin{equation*}
\eta_t=\eta_{\max}\min\left\{
\frac{n_{\mathrm{KRR}}(t)}{3},\,1
\right\},
\end{equation*}
where $n_{\mathrm{KRR}}(t)$ denotes the number of completed KRR fits and
$\eta_{\max}$ is the environment-specific coefficient given in
Table~\ref{tab:app_config}.
For a task reward $r_t\in\{0,1\}$, the execution label and residual target are
\begin{equation*}
y_t=2r_t-1,\qquad
\widehat{y}^{\mathrm{base}}_t=2\sigma\!\left(
 s_{\mathrm{base},t}(q_t,m_t^\star)\right)-1,\qquad
\delta_t=y_t-\widehat{y}^{\mathrm{base}}_t.
\end{equation*}
Only the selected pair $(q_t,m_t^\star)$ receives feedback. These labels
reflect downstream task outcomes, not same-task comparisons among memories.

\paragraph{Kernel ridge regression and AGOP.}
The residual predictor fits kernel ridge regression (KRR) to a FIFO window
of at most $128$ feedback pairs $\{(\mathbf{z}_j,\delta_j)\}$, using the
Laplacian kernel
\begin{equation*}
K_{\mathbf{M}_t}(\mathbf{z},\mathbf{z}')=
\exp\!\left[-\gamma\sqrt{(\mathbf{z}-\mathbf{z}')^\top
\mathbf{M}_t(\mathbf{z}-\mathbf{z}')}\right],
\end{equation*}
with $\lambda_{\mathrm{KRR}}=0.01$ and $\gamma=1.0$. The metric
$\mathbf{M}_t$ is diagonal. Fast KRR coefficients are periodically refitted;
slower feature-geometry updates use the diagonal Average Gradient Outer
Product (AGOP):
\begin{align*}
\widetilde{\mathbf{G}}_t
&=\operatorname{Diag}\!\left(\frac{1}{N_s}\sum_{j=1}^{N_s}
\nabla_{\mathbf{z}}f_t(\mathbf{z}_j)\odot
\nabla_{\mathbf{z}}f_t(\mathbf{z}_j)\right),\\
\mathbf{M}_{t+1}
&=\operatorname{NormClip}\!\left((1-\beta_M)\mathbf{M}_t+
\beta_M\widetilde{\mathbf{G}}_t\right).
\end{align*}
The slow window contains at most $256$ feedback examples. We set
$\beta_M=0.25$, clip diagonal weights to $[0.05,8.0]$, and normalize them to
mean $1$. KRR is refitted after every AGOP update; update intervals are
listed in Table~\ref{tab:app_config}.

\paragraph{Execution-derived memory statistics.}
A newly created memory starts with utility $u(m)=1.0$. After a task with
selected memory $m_t^\star$, its utility is updated without an EMA:
\begin{equation*}
u_t(m_t^\star)=\operatorname{clip}\!\left(
 u_{t-1}(m_t^\star)+
 \begin{cases}+0.25,&r_t=1,\\-0.45,&r_t=0,\end{cases}
 -2,5\right).
\end{equation*}
Successful (unsuccessful) task execution with the selected memory increments
$h_m$ ($d_m$). Access count and recency are updated after each use.

\paragraph{Memory admission, replacement, and eviction.}
Only externally verified successful tasks produce candidate memories.
While $|\mathcal{B}_t|<B_{\max}$, each eligible candidate is admitted to
establish coverage. For a full bank, let $m_{\mathrm{near}}$ denote the
nearest existing memory in the same hashed textual representation. Define
\begin{align*}
\operatorname{nov}_t(m_{\mathrm{new}})
&=1-\max_{m_i\in\mathcal{B}_t}
  \operatorname{cos}(\mathbf{e}_{\mathrm{new}},\mathbf{e}_i),\\
V_t(m)
&=\frac{1}{|\mathcal{P}_t|}
  \sum_{q\in\mathcal{P}_t}s_t(q,m),\\
\rho_t^+(m_{\mathrm{new}})
&=\frac{1}{|\mathcal{P}_t|}
  \sum_{q\in\mathcal{P}_t}
  \mathbb{I}\!\left[s_t(q,m_{\mathrm{new}})>0\right].
\end{align*}
The probe set $\mathcal{P}_t$ includes the current query and up to eight
previous queries on which memory retrieval occurred (FIFO), hence
$|\mathcal{P}_t|\leq9$. The positive-score ratio $\rho_t^+$ measures the
fraction of probe queries for which the candidate receives a positive worth
score, rather than the fraction on which it outperforms the nearest memory.

For a full bank, a candidate passes the admission gate if
\begin{equation*}
\begin{aligned}
&\operatorname{nov}_t(m_{\mathrm{new}})>\tau_{\mathrm{nov}}\\
&\quad\lor\left[
V_t(m_{\mathrm{new}})
>\max_{m\in\mathcal{B}_t}V_t(m)+\delta_{\mathrm{write}}
\;\land\;
\rho_t^+(m_{\mathrm{new}})\geq\tau_{\mathrm{pos}}
\right],
\end{aligned}
\end{equation*}
where $\tau_{\mathrm{nov}}=0.35$, $\delta_{\mathrm{write}}=0.02$,
and $\tau_{\mathrm{pos}}=0.45$. Thus, admission requires either sufficient
novelty or a worth improvement over the best existing memory together with
a sufficiently high positive-score ratio.

Candidates passing this admission gate are added to the bank. The capacity
bound is then enforced by evicting the memory with the lowest probe-averaged
worth from the enlarged bank:
\begin{equation*}
m_{\mathrm{evict}}
=\arg\min_{m\in\mathcal{B}_t\cup\{m_{\mathrm{new}}\}}V_t(m),
\qquad |\mathcal{B}_{t+1}|\leq B_{\max}.
\end{equation*}
The newly admitted candidate may itself be evicted if it has the lowest
probe-averaged worth.

If the candidate fails the admission gate, the controller instead considers
direct replacement of its nearest existing memory. Replacement occurs only
when the nearest memory has low utility and the candidate exceeds its
probe-averaged worth by a larger margin:
\begin{equation*}
u(m_{\mathrm{near}})<\tau_{\mathrm{util}}=0.5,
\qquad
V_t(m_{\mathrm{new}})
>V_t(m_{\mathrm{near}})+\delta_{\mathrm{rep}},
\quad \delta_{\mathrm{rep}}=0.10.
\end{equation*}
When both conditions hold, $m_{\mathrm{new}}$ directly replaces
$m_{\mathrm{near}}$ without invoking the lowest-worth eviction rule.
Candidates satisfying neither the admission gate nor the fallback
replacement conditions are rejected.

\subsection{Common Hyperparameters}
The main common parameters are $\mathbf{w}_0=[1.0,0.10,0.10,0.02]$,
$\lambda_w=0.1$, $\beta_w=0.5$, $\lambda_{\mathrm{KRR}}=0.01$,
$\gamma=1.0$, $\beta_M=0.25$, and $\kappa=0.02$. The base, fast KRR,
and slow AGOP feedback windows contain at most $256$, $128$, and $256$
examples, respectively. The shared memory-management thresholds are
$\tau_{\mathrm{nov}}=0.35$, $\delta_{\mathrm{write}}=0.02$,
$\tau_{\mathrm{pos}}=0.45$, and $\delta_{\mathrm{rep}}=0.10$.
Further bounds and update rules are specified in Appendix~\ref{app:implementation}
above, and the environment-dependent settings in Table~\ref{tab:app_config}.

\subsection{Environment-Specific Configuration}
TokenWorth uses the same worth-learning method across DB, OS, and KG, with
resource limits and online-update intervals adapted to the environment.
\begin{table}[!htbp]
\centering
\caption{Environment-specific resource and adaptation configurations for LifelongAgentBench.}
\label{tab:app_config}
\small
\setlength{\tabcolsep}{9pt}
\renewcommand{\arraystretch}{1.06}
\begin{tabular}{@{}lccc@{}}
\toprule
Setting & DB & OS & KG \\
\midrule
Generation budget $O_{\max}$ & 128 & 512 & 128 \\
Interaction budget $J_{\max}$ & 3 & 5 & 15 \\
Memory capacity $B_{\max}$ & 64 & 64 & 64 \\
Retrieved memories $K_{\max}$ & 1 & 1 & 1 \\
Injection budget $L_{\max}$ & 900 chars & 800 chars & 4096 chars \\
Candidate pool size $K_c$ & 12 & 12 & 12 \\
KRR update interval & 50 & 25 & 25 \\
AGOP update interval & 200 & 100 & 100 \\
Residual influence $\eta$ & 0.15 & 0.05 & 0.10 \\
RFM warm-up feedback & 4 & 4 & 24 \\
\bottomrule
\end{tabular}
\end{table}
\FloatBarrier

\subsection{Environment-Specific Memory Construction}

\paragraph{DB.}
Successful tasks are stored as compact demonstrations containing the
question, valid SQL actions, truncated database observations, and final
answer. Invalid actions and unnecessary output are omitted.

\paragraph{OS.}
Complete shell trajectories contain instance-specific paths, user names,
and process identifiers that transfer poorly. We retain concise patterns
from successful Bash actions, including permission management, batch file
processing, text editing, symbolic links, and process management. A task is
successful when the environment evaluator reports \texttt{correct}, including
when the interaction limit is reached after a correct final state.

\paragraph{KG.}
Memories contain successful knowledge-graph tool calls and truncated
knowledge-graph observations, rather than SPARQL queries executed by the
backend. KG allows up to $15$ interaction rounds and $4096$ characters
of injected memory; residual adaptation starts after at least $24$
query--memory feedback pairs. In all three environments, TokenWorth first
screens up to $K_c=12$ memories and injects at most one selected experience.

\section{Generalization Beyond LifelongAgentBench}
\label{app:generalization}

We evaluate online lifelong variants of
BigCodeBench~\citep{zhuo2025bigcodebench}
(code generation) and ALFWorld~\citep{shridhar2020alfworld}
(text-based embodied decision making). Each run starts without stored
experience from benchmark tasks. Tasks are processed sequentially, with
retrieval using only preceding experience. Memory is updated after the
current task has been evaluated, without access to future tasks.
The language-model backbone remains frozen throughout each run.

\paragraph{BigCodeBench.}
We use BigCodeBench Full v0.1.4 in the \textbf{Instruct} format,
comprising \textbf{1,140} coding problems. Each method makes a single
pass through the same deterministic sequence: the 992 problems in
Full excluding Hard, followed by the 148 problems in the Hard subset.
Within these two groups, we preserve the order of the pinned Full and
Hard dataset files, respectively; no random shuffling is applied.
The memory bank is carried across the two groups without resetting.
For each problem, the agent generates one solution, which is sanitized
and evaluated using BigCodeBench's official execution-based evaluator
inside a Docker worker. A problem succeeds only when the evaluator
returns \texttt{pass}; generated-code failures and execution timeouts
count as unsuccessful outcomes. Evaluator infrastructure errors
invalidate the run. We report pass@1 over all 1,140 problems, counting
each problem once. Memory is updated after evaluation, and previously
evaluated problems are not revisited. The local deployment uses
Qwen2.5-14B-Instruct. The API deployment uses DeepSeek-V4-Flash with thinking explicitly disabled.
Both deployments use temperature zero. Reference solutions and unit
tests are excluded from the model's memory and input prompts.

\paragraph{ALFWorld.}
We use Qwen2.5-14B-Instruct as the frozen backbone, with temperature
zero and top-$p$ equal to one. A complete run contains \textbf{574}
episodes: 300 training episodes selected by task-type-stratified
sampling, all 140 validation-seen episodes, and all 134
validation-unseen episodes. The phases run in this order. The base
seed is 42; the sampling and ordering routines use seeds 42, 43, and
44 for the three phases, respectively. Within each phase, game paths
are sorted, shuffled within task-type groups using the phase seed,
and selected by round-robin traversal of the groups. The resulting
ordered episode manifest is fixed and shared across methods, and
the runner checks that the environment follows this order.
Each selected episode is executed once. An episode succeeds when
TextWorld's external \texttt{won} flag indicates task completion.
The success rate reported here is the total number of successful
episodes divided by \textbf{574}, including all three phases; it is
not an unweighted average of phase success rates. Memory persists
across phase boundaries and is updated after every completed episode,
including validation-seen and validation-unseen episodes. Thus, this
is an online lifelong protocol rather than evaluation with memory
frozen after training. Expert trajectories are excluded from memory.

\paragraph{Resource accounting.}
We follow the resource-constrained evaluation principle in Sec.~3,
using benchmark-specific limits. Within each benchmark and deployment,
methods share the backbone, ordered task stream, solver-call output
limit, and, for ALFWorld, interaction controller. Memory contexts are
bounded by the complete injected block, including its headings and
separators. Injection lengths below are measured in \emph{characters}.
The two adapters enforce output limits \emph{per model call}, rather
than pooling all calls into a single per-task output allowance.
Method-specific auxiliary calls, including memory construction,
reflection, and skill extraction, have a separate limit of 512 output
tokens per call. Their actual usage is recorded in addition to solver
usage, including action repairs in ALFWorld. TokenWorth makes no
auxiliary language-model calls. Consequently, the methods share
per-call limits but need not consume equal total output tokens per
task or episode.

\paragraph{BigCodeBench configuration.}
The solver output limit is $O_{\max}=1280$ tokens per call for both the local and API deployments.
Auxiliary API calls have thinking disabled.
TokenWorth uses a memory capacity of $B_{\max}=50$ records, injects
at most $K_{\max}=1$ record, and caps the complete injection block at
$L_{\max}=4096$ characters. Each stored experience is also capped at
4096 characters. BM25 recalls at most $K_c=12$ candidates for learned
worth ranking. All TokenWorth representation, learning, and
memory-management parameters are identical across the local and
API deployments; the solver output budget and thinking setting
follow the selected deployment.

\paragraph{ALFWorld configuration.}
The interaction limit is $J_{\max}=50$ environment steps per episode.
Each action-generation call has an output limit of $O_{\max}=128$
tokens. An inadmissible action permits at most one repair call for
that step, also capped at 128 output tokens. These are per-call
limits, not a 128-token allowance for the entire episode. The same
action controller is used for all methods, with memory-dependent
loop removal disabled. TokenWorth uses $B_{\max}=128$ records,
$K_{\max}=1$, a complete injection budget of $L_{\max}=4200$
characters, a per-experience limit of 4096 characters, and a BM25
candidate pool of at most $K_c=12$.

\paragraph{Online learning configuration.}
Both benchmarks use the same deterministic 512-dimensional hashing
representation of tokens and character 3--5-grams. Only a memory
actually injected into the agent receives outcome supervision.
KRR and AGOP updates are scheduled every 75 and 225 accumulated
feedback examples, respectively, with FIFO buffers of 128 and 256
examples. These counters track feedback examples, not elapsed tasks
or episodes. Each scheduled residual fit requires at least four
feedback examples; this minimum does not trigger an immediate fit
at the fourth example. Under these settings, the first scheduled
KRR fit occurs at feedback example 75. The residual coefficient is
$\eta_t=0.10\min(1,u_t/3)$, where $u_t$ is the number of completed
KRR fits: its influence ramps linearly over the first three fits.
KRR uses ridge regularization $0.01$ and Laplacian-kernel scale $1$.
AGOP uses smoothing $0.25$, clips the updated diagonal metric to
$[0.05,8]$, normalizes it to have mean one, and immediately refits
KRR. The base scorer is refitted every 50 feedback examples, using
a FIFO of 256, regularization $0.1$, and smoothing $0.5$.
Successful interactions can create new experiences; unsuccessful
interactions provide negative feedback to injected memories without
creating successful-experience records.

\begin{table}[!htbp]
\centering
\caption{Task success rate (\%) on additional generalization benchmarks.
Best results are \textbf{bold}; second-best results are \underline{underlined}
(including ties).}
\label{tab:generalization}
\small
\setlength{\tabcolsep}{7pt}
\renewcommand{\arraystretch}{1.06}
\begin{tabular}{@{}lccc@{}}
\toprule
& \multicolumn{2}{c}{\textit{Local: Qwen2.5-14B-Instruct}}
& \textit{API: DeepSeek-V4-Flash} \\
\cmidrule(lr){2-3}\cmidrule(l){4-4}
Method & \textsc{BigCode} & \textsc{ALFWorld} & \textsc{BigCode} \\
\midrule
Standard & 41.7 & 59.1 & 49.9 \\
RAG & 42.6 & \underline{73.2} & \textbf{50.9} \\
MemRL & \underline{42.9} & 69.7 & 50.5 \\
Mem0 & 41.7 & 64.1 & 50.2 \\
MemP & 42.3 & 66.4 & 49.6 \\
MemSkill & \underline{42.9} & 63.6 & 49.1 \\
Reflexion & 40.8 & 69.2 & 49.0 \\
EvoSC & 42.6 & 65.0 & 49.6 \\
EvoSkill & 41.8 & 64.8 & 48.8 \\
\textbf{TokenWorth} & \textbf{43.1} & \textbf{73.5} & \underline{50.6} \\
\bottomrule
\end{tabular}
\end{table}
\FloatBarrier

Table~\ref{tab:generalization} shows that TokenWorth achieves $43.1\%$
on local BigCodeBench and $73.5\%$ on local ALFWorld, exceeding the best
competitors by $0.2$ and $0.3$ percentage points, respectively. On API-based
BigCodeBench, TokenWorth reaches $50.6\%$, below RAG's $50.9\%$. These
small margins support competitive transfer rather than consistent dominance
beyond the three main interactive environments.

\section{Hyperparameter Sensitivity}
\label{app:hyperparameter_sensitivity}

We evaluate the sensitivity of TokenWorth to its key hyperparameters on
the same $500$-task DB stream using Qwen2.5-14B-Instruct. Starting from
the default configuration, which achieves a $65.0\%$ task success rate,
we vary one parameter at a time while keeping all others fixed.
Table~\ref{tab:hyperparameter_sensitivity} summarizes the results.

\begin{table}[!htbp]
\centering
\caption{Hyperparameter sensitivity on DB using Qwen2.5-14B-Instruct.
The default configuration achieves $65.0\%$ task success.
$\Delta$ denotes the change in percentage points relative to the default.}
\label{tab:hyperparameter_sensitivity}
\small
\setlength{\tabcolsep}{8pt}
\renewcommand{\arraystretch}{1.06}
\begin{tabular}{@{}lcccc@{}}
\toprule
Parameter & Default & Tested value & SR (\%) & $\Delta$ (pp) \\
\midrule
\multirow{4}{*}{RFM influence $\eta$}
& \multirow{4}{*}{0.15} & 0.05 & 62.4 & $-2.6$ \\
& & 0.10 & 63.8 & $-1.2$ \\
& & 0.20 & 64.2 & $-0.8$ \\
& & 0.30 & 64.0 & $-1.0$ \\
\midrule
\multirow{3}{*}{Base regularization $\lambda_w$}
& \multirow{3}{*}{0.10} & 0.03 & 64.8 & $-0.2$ \\
& & 0.30 & 62.6 & $-2.4$ \\
& & 1.00 & 63.0 & $-2.0$ \\
\midrule
\multirow{3}{*}{Base smoothing}
& \multirow{3}{*}{0.50} & 0.25 & 64.2 & $-0.8$ \\
& & 0.75 & 62.8 & $-2.2$ \\
& & 1.00 & 62.6 & $-2.4$ \\
\midrule
\multirow{2}{*}{Base update interval}
& \multirow{2}{*}{50} & 25 & 65.2 & $+0.2$ \\
& & 100 & 62.4 & $-2.6$ \\
\midrule
\multirow{3}{*}{AGOP smoothing}
& \multirow{3}{*}{0.25} & 0.10 & 64.4 & $-0.6$ \\
& & 0.50 & 64.6 & $-0.4$ \\
& & 0.75 & 62.2 & $-2.8$ \\
\midrule
\multirow{2}{*}{KRR update interval}
& \multirow{2}{*}{50} & 25 & 62.2 & $-2.8$ \\
& & 100 & 63.4 & $-1.6$ \\
\midrule
\multirow{2}{*}{AGOP update interval}
& \multirow{2}{*}{200} & 100 & 62.6 & $-2.4$ \\
& & 400 & 62.0 & $-3.0$ \\
\midrule
\multirow{2}{*}{Kernel $\gamma$}
& \multirow{2}{*}{1.0} & 0.5 & 64.0 & $-1.0$ \\
& & 2.0 & 63.0 & $-2.0$ \\
\bottomrule
\end{tabular}
\end{table}
\FloatBarrier

TokenWorth maintains relatively stable performance under moderate
changes to several hyperparameters. Varying the RFM influence $\eta$
between $0.10$ and $0.30$ results in performance differences of at most
$1.2$ percentage points, while adjusting AGOP smoothing between $0.10$
and $0.50$ leads to differences within $0.6$ points. Reducing the base
update interval from $50$ to $25$ yields a comparable success rate
of $65.2\%$.

Larger deviations in adaptation settings lead to more pronounced
performance degradation. Reducing $\eta$ to $0.05$ decreases success by
$2.6$ points, while increasing AGOP smoothing to $0.75$ causes a
$2.8$-point drop. The update schedules also play an important role:
changing the KRR update interval from $50$ to $25$ reduces success by
$2.8$ points, while changing the AGOP interval from $200$ to $100$ or
$400$ decreases success by $2.4$ and $3.0$ points, respectively.
These results highlight the importance of balancing adaptation speed
and stability, supporting the conservative, two-timescale design
of TokenWorth.

\end{document}